\pdfoutput=1

\documentclass[11pt]{article}

\usepackage[]{acl}

\usepackage{times}
\usepackage{latexsym}

\usepackage[T1]{fontenc}

\usepackage[utf8]{inputenc}

\usepackage{microtype}

\usepackage{times}
\usepackage{graphicx}
\usepackage{caption}
\usepackage{subcaption}
\usepackage{latexsym}
\usepackage{amsmath}

\usepackage{amsmath}
\usepackage{amssymb}
\usepackage{enumitem}
\usepackage{multirow}
\usepackage{enumitem}
\usepackage{makecell}
\usepackage{tablefootnote}
\usepackage[utf8]{inputenc}
\usepackage{comment}
\usepackage[english]{babel}
\usepackage{xcolor}
\usepackage{float}

\title{Improving Generalizability in Implicitly Abusive Language Detection with Concept Activation Vectors}

 \author{Isar Nejadgholi, Kathleen C. Fraser, and Svetlana Kiritchenko \\
  National Research Council Canada \\
  Ottawa, Canada \\
 \footnotesize \texttt{\{Isar.Nejadgholi,Kathleen.Fraser,Svetlana.Kiritchenko\}@nrc-cnrc.gc.ca}\\
 }

\begin{document}
\maketitle
\begin{abstract}
Robustness of machine learning models on ever-changing real-world data is critical, especially for applications affecting human well-being such as content moderation.  
New kinds of abusive language continually emerge in online discussions in response to current events (e.g., COVID-19), and the deployed abuse detection systems should be 
updated regularly to remain accurate.
In this paper, we show that general abusive language classifiers tend to be fairly reliable in detecting out-of-domain explicitly abusive utterances but fail to detect new types of more subtle, implicit abuse. 
Next, we propose an interpretability technique, based on the Testing Concept Activation Vector (TCAV) method from computer vision, to quantify the sensitivity of a trained model to the human-defined concepts of explicit and implicit abusive language, and use that to explain the generalizability of the model on new data, in this case, COVID-related anti-Asian hate speech. Extending this technique, we introduce a novel metric, \textit{Degree of Explicitness}, for a single instance and show that the new metric is beneficial in suggesting out-of-domain unlabeled examples to effectively enrich the training data with informative, implicitly abusive texts. 
\end{abstract}

\section{Introduction}

When machine learning models are deployed in the real world, they must be constantly monitored for their robustness to new and changing input data. One area where this is particularly important is in abusive language detection \cite{schmidt2017survey,fortuna2018survey,nakov2021detecting,vidgen2020directions}. The content of online conversation is constantly changing in response to political and social events. New categories of abusive language emerge, encompassing topics and vocabularies unknown to previously trained classifiers. Here, we tackle three main questions: How can a human user formalize new, relevant topics or concepts in text? How do we quantify the sensitivity of a trained classifier to these new concepts as they emerge? 
And 
how do we update the classifier so that it remains reliable?

As a case study, we consider the rise of COVID-related anti-Asian racism on social media. The COVID-19 pandemic represented an entirely new and unexpected situation, generating new vocabulary (\textit{COVID-19}, \textit{coronavirus}, \textit{social distancing}, \textit{masking}), new topics of conversation (dealing with isolation, working from home), and -- unfortunately -- new and renewed instances of hate speech directed towards Asian communities. We imagine the case of an abusive language detection algorithm which had been deployed prior to the pandemic: 
what are the new types of abusive language that have emerged with the recent pandemic? To what extent can deployed classifiers generalize to this new data, and how can they be adapted?
Although social events can spark off a specific type of hate speech, they are rarely the root cause of the issue. Often such hateful beliefs existed before the event, and are only magnified because of it \cite{chou2015myth}. Therefore, we expect that the classifier should detect this new variety of hate speech to some extent.

An important factor in this study is whether the text expresses explicit or implicit abuse \cite{waseem-etal-2017-understanding,caselli-etal-2020-feel,wiegand-etal-2021-implicitly}. Explicit abuse refers to utterances that include direct insults or strong rudeness, often involving profanities, whereas implicit abuse involves more indirect and nuanced language. Since understanding the offensive aspects of implicit abuse in our case study may require some knowledge of the context (i.e., the pandemic), we expect that the pretrained classifier will find these data especially difficult to handle.

To examine a classifier's ability to handle new type of abusive text (without access to extensive labeled data), we propose a technique based on the Testing Concept Activation Vector (TCAV) method from the interpretability literature in computer vision \cite{kim2018interpretability}. TCAV is used to explain whether a classifier associates a specific concept to a class label (e.g., the concept of \textit{stripes} is associated with class \textit{zebra} in image classification). 
Similarly, we define implicit and explicit COVID-related anti-Asian racism with a small set of human-chosen textual examples, and ask whether the pretrained classifier associates these concepts with the positive (abusive) class label.

Further, we ask whether sensitivity to human-defined concepts can direct data augmentation\footnote{In this paper, we use the term \textit{augmentation} to refer to the process of enriching the training data by adding examples from sources other than the original dataset.} to improve generalizations. Intuitively, when updating a classifier, data enrichment should focus on adding examples of concepts to which the classifier is not yet sensitive. Conventional active learning frameworks suggest examples with the lowest classification confidence as the most informative augmentation samples \cite{zhu2008learning,chen2019active}. However, deep neural networks' inability to provide reliable uncertainty estimates is one of the main barriers to adopting confidence-based sampling techniques \cite{schroder2020survey}. We suggest that, in the case of abuse detection, implicitly abusive examples are most informative for updating a general classifier. However, to the best of our knowledge, there is no quantitative metric that can measure the degree of explicitness of a candidate example, given a trained classifier. We extend the TCAV technique to provide a “degree of explicitness” measure at the utterance level and use that for efficient data augmentation. 

The contributions of this work are as follows:
\vspace{5pt}
\setlist{nolistsep}
\begin{itemize}[noitemsep,leftmargin=10pt]

\item We implement a variation of the TCAV framework for a RoBERTa-based classifier and show that it can be used to quantify the sensitivity of a trained classifier to a human-understandable concept, defined through examples, without access to the training dataset of the classifier or a large annotated dataset for the new category.

\vspace{5pt}
\item We analyse the performance of two abusive language classifiers and observe that they generalize well to explicit COVID-related anti-Asian racism, but are unable to generalize to implicit racism of this type. We show that sensitivities to the concepts of implicit and explicit abuse can explain the observed discrepancies. 

\vspace{5pt}
\item We adjust the TCAV method to compute the \textit{degree of explicitness}, for an unlabeled instance, as a metric to guide data augmentation when updating a general abusive language classifier to include a new kind of abuse. We test this method against confidence-based augmentation and show that it is able to reach higher accuracy 
with fewer training examples,
while maintaining the accuracy on the original data.
\end{itemize}

\noindent The implementation code and data for the experiments are available at \url{https://github.com/IsarNejad/TCAV-for-Text-Classifiers}.

\section{Datasets and Data Analysis}

We consider the following four English datasets, summarized in Table~\ref{tab:data}: \textit{Founta}\footnote{For \textit{Founta}, we discard the tweets labeled as Spam and use the train-dev-test split as provided by \citet{zhou2021challenges}.} and 
\textit{Wiki}\footnote{We used a smaller version of the \textit{Wiki} dataset as provided in \citet{nejadgholi-kiritchenko-2020-cross}. In that work, we removed 54\% of Wikipedia-specific non-toxic instances from the training set to mitigate the topic bias, and reported improvements in both the classification performance and the execution time. Here, we found similar benefits.}
are large, commonly-used datasets for general abusive language detection, while \textit{EA} and \textit{CH} specifically target COVID-related anti-Asian racism. 
 We binarize all datasets to two classes: positive (i.e., abusive or hateful) and negative. For \textit{Founta}, this means combining Abusive and Hateful texts into a single positive class; for \textit{EA}, ``Hostility against an East-Asian entity'' is considered positive, and all other classes are negative; and for \textit{CH}, all hate speech is classed as positive, while counter-hate and hate-neutral texts are classed as negative.

\setlength{\tabcolsep}{5pt}
\begin{table*}[tbh]
    \centering
    \small
    \begin{tabular}{ p{2.8cm} p{1.2cm} p{3.3cm} p{3.3cm} r r r }
    \hline
    \textbf{Dataset} & \textbf{Data} & \textbf{Positive Class }& \textbf{Negative Class} & \multicolumn{3}{c}{\textbf{Number (\%Pos:\%Neg)}} \\
    & \textbf{Source} & & & \textit{Train} & \textit{Dev} & \textit{Test} \\
    \hline
 \makecell[tl]{Wikipedia\\ Toxicity (\textit{Wiki}) \\\cite{wulczyn2017ex}} & \makecell[tl]{Wikipedia\\comments} & Toxic & Normal &\makecell{43,737\\(17:83)} &\makecell{32,128\\(9:91)} & \makecell{31,866\\(9:91)} \\
 \hline
  \makecell[tl]{\citet{Founta2018}\\ dataset (\textit{Founta})} & \makecell[tl]{Twitter\\posts} &  Abusive; Hateful &  Normal & \makecell{62,103\\(37:63)} & \makecell{10,970\\(37:63)} &\makecell{12,893\\(37:63)}\\
 \hline
 \makecell[tl]{East-Asian\\ Prejudice (\textit{EA})\\ \cite{vidgen2020detecting}}  & \makecell[tl]{Twitter\\posts} &  \makecell[tl]{Hostility against an\\ East Asian entity} & \makecell[tl]{Criticism of an East Asian\\ entity; Counter speech;\\ Discussion of East Asian\\ prejudice; Non-related} &\makecell{16,000\\(19:81)} &\makecell{1,200\\(19:81)} &\makecell{2,800\\(19:81)}\\
 \hline
 \makecell[tl]{COVID-HATE (\textit{CH})\\ \cite{ziems2020racism}} & \makecell[tl]{Twitter\\posts} & \makecell[tl]{Anti-Asian COVID-19\\ hate; 
 Hate directed to\\ non-Asians} & \makecell[tl]{Pro-Asian COVID-19\\ counterhate;\\ Hate-neutral} & -- & --  & \makecell{2,319\\(43:57)} \\
 
\hline
    \end{tabular}
    \caption{ Class descriptions, number of instances and ratio of positive to negative in percentage (\%Pos:\%Neg) for the general abusive datasets (\textit{Wiki} and \textit{Founta}) and COVID-related Anti-Asian hate speech datasets (\textit{EA} and \textit{CH}).}
    \label{tab:data}
\end{table*}
\setlength{\tabcolsep}{6pt}

\subsection{Differences in Vocabulary}
Central to our research question is the issue of vocabulary change as a new abusive topic emerges. As the \textit{Wiki} and \textit{Founta} datasets were collected before the COVID-19 pandemic, they do not contain novel vocabulary such as ``chinavirus'' or ``wuhanflu'', and the contexts and frequencies for words like ``China'' and ``pandemic'' may have changed. As a demonstration of the differences in vocabulary across the different datasets, we compute the top 100
most frequent words in the positive class of each dataset (after removing stop words\footnote{We used the stop word list from the scikitlearn package.}), and then calculate the overlap between each pair of datasets. 
We categorize the shared words into three categories: 1) generically profane and hateful words, 2) COVID-related words, and 3) all other words.  
Table \ref{tab:shared-words} shows the three categories of shared words among the 100 most frequent words of the positive classes in our datasets. 

 
This analysis reveals that the two COVID-related datasets share more words in common: 50 out of the 100 most frequent words are common between the two datasets. As expected, a large portion of their shared vocabulary (32\%) is specific to the pandemic, has been used more frequently during the pandemic or has found new connotations because of the pandemic. For all other datasets, fewer words are shared, and the shared words are either related to profanity and violence or are merely commonly used terms.  
Profanity and strongly negative words such as ``hate'' make up 30\% of the shared vocabulary between the \textit{Wiki} and \textit{Founta} datasets. Interestingly, \textit{CH} has a set of profane words in common with both \textit{Wiki} and \textit{Founta} ($\sim$25\% of shared words), while the words shared between \textit{EA} and the general datasets are simply common words in the English language, such as ``people'', ``want'', and ``need.'' We expect that this vocabulary shift between the different datasets will have a considerable impact on the generalizability.

\setlength{\tabcolsep}{4pt}
\begin{table*}[hbt!]
    \centering
    \small
    \begin{tabular}{ p{1.7cm} c p{11.5cm}}
    \hline
    \textbf{Datasets} & \textbf{Count} & \textbf{Shared Words}\\
    \hline
     EA - CH&50&\textbf{COVID-related (32\%):} ccp, 19, communist, pandemic, coronavirus,  covid19, chinesevirus, infected, covid,  chinese, chinavirus, corona, wuhanvirus, wuhan, china, virus
     \\
     &&\textbf{Hateful (0\%)}\\
     &&\textbf{Other (68\%):} racist, came, want, country, calling, come, does, spread, like, amp, media, eating, did, human, world, know, government, say, started, think, need, blame, evil, time, people, don, new, let, news, stop, countries, just, spreading, make\\
     \hline
    Wiki - Founta & 37&\textbf{COVID-related (0\%)}\\
    &&\textbf{Hateful (30\%):} *ss, b*tch, id*ot, n*ggas, d*ck, f*cking, f*ck,  sh*t,  hell, hate, stupid\\
    &&\textbf{Other (70\%):} oh, dont, want, way, going, come, does, like, look, life, did eat, sex, know, say, think, man, need, time, people, said, stop, really, just, make, tell\\
    \hline
    Founta - EA&19&\textbf{COVID-related (0\%)}\\
    &&\textbf{Hateful (0\%)} \\
    &&\textbf{Other (100\%):} racist, want, calling, come, does, like, did, world, know, say, think, need, time, people, trying, let, stop, just, make\\
    \hline
Wiki - EA&15&\textbf{COVID-related (0\%)}\\
    &&\textbf{Hateful (0\%)} \\
    &&\textbf{Other (100\%):} people, want, did, say, think, good, need, come, does, stop, just, know, like, make, time\\
    \hline
Founta - CH&35&\textbf{COVID-related (0\%)}\\
    &&\textbf{Hateful (23\%):} *ss, b*tch, f*cking, f*ck, sh*t, hate, stupid, f*cked\\
    &&\textbf{Other (77\%):} racist, want, way, going, calling, come, does, like, got, look, did, eat, world, know, say, think, man, trump, need, time, people, said, let, stop, really, just, make\\
    \hline
Wiki - CH&33&\textbf{COVID-related (0\%)}\\
    &&\textbf{Hateful (27\%):} *ss, b*tch, f*cking, f*ck,  sh*t, hate, stupid, shut,  kill\\
    &&\textbf{Other (73\%):} want, way, going, come, does, like, look, did, eat, right, know, die, say, think, man, need, time, people, don, said, stop, really, just, make\\
    \hline
    \end{tabular}
    \caption{Shared words among 100 most frequent words of the positive classes in the datasets.}
    \label{tab:shared-words}
\end{table*}
\setlength{\tabcolsep}{4pt}

\subsection{Differences in Explicitness}
Another important factor in our study is generalization with respect to explicit and implicit types of abusive language. 
Above, 
we observed that \textit{CH} shares many profane words with the general datasets 
and, therefore, we anticipate it contains more explicitly abusive texts than \textit{EA} does. 
Unfortunately, neither of the datasets has originally been annotated for \textit{explicitness of abuse}. 
We manually annotate instances from the positive class in the \textit{CH} dataset and the \textit{EA} dev set 
using the following rule: 
instances that include profanity, insult or rudeness 
that could be correctly identified as abusive without general knowledge about the COVID-19 pandemic are labeled as explicitly abusive; the remaining instances  (e.g., \textit{`it is not covid 19 but wuhanvirus'}) are labeled as implicitly abusive. We find that 85\% of the \textit{CH-positive} class is categorized as explicit, whereas only 8\% of the \textit{EA-positive} class in the \textit{EA} dev set is labeled as explicit. 
Thus, \textit{CH} and \textit{EA} share COVID-related vocabulary, but are very different in terms of explicitness of abuse (\textit{CH} containing mostly explicit abuse while \textit{EA} containing mostly implicit abuse), which makes them suitable test beds for assessing the generalizability of classifiers to a new type of abusive language and the impact of new vocabulary on the classification of implicit and explicit abuse.

\section{Cross-Dataset Generalization}
\label{sec:generalization}

We start by assessing the robustness of a general-purpose abusive language classifier on a new domain of abusive language. Specifically, we analyze the performance of classifiers trained on the \textit{Wiki} and \textit{Founta} datasets (expected to detect general toxicity and abuse) on COVID-related anti-Asian racism data. 
In addition, we want to assess the impact of the change of vocabulary on the generalizibility of the 
classifiers to implicit and explicit 
abuse in the new domain. 
We train binary RoBERTa-based classifiers on the \textit{Wiki}, \textit{Founta}, \textit{EA} and \textit{CH} datasets (referred to hereafter as the \textit{Wiki}, \textit{Founta}, \textit{EA} and \textit{CH} classifiers), and test them on the \textit{EA} as the mostly implicit COVID-related dataset and \textit{CH} as the mostly explicit COVID-related dataset. (The training details are provided in Appendix \ref{sec:appendix-A}.) Note that \textit{CH} is 
too small to be broken into train/test/dev sets, so it is used either as a training dataset when testing on \textit{EA} or a test dataset for all other classifiers. Here, while the classifier makes a binary positive/negative decision, we are really assessing its ability to generalize to the new task of identifying anti-Asian hate. 
For comparison, we also train an ``explicit general abuse'' classifier with only explicit examples of the \textit{Wiki} dataset and the class balance similar to the original \textit{Wiki} dataset. This classifier is referred to as \textit{Wiki-exp}.\footnote{For \textit{Wiki-exp}, the examples of the positive class are taken from the `explicit abuse' topic, which contains texts with explicitly toxic words, from \cite{nejadgholi-kiritchenko-2020-cross}, and negative examples are randomly sampled from the \textit{Wiki-Normal} class.}

Table~\ref{tab:generalization} presents the Area Under the ROC Curve (AUC) and F1-scores for all the classifiers; precision, recall, and average precision scores are provided in Appendix \ref{sec:appendix-A1}. 
We first consider whether class imbalances can explain our results. Note that while abusive language is a relatively rare phenomenon in online communications, most abusive language datasets are collected through boosted sampling and therefore are not subject to extreme class imbalances. The percentage of positive instances in our datasets ranges from 9\% to 43\% (Table~\ref{tab:data}). 
We observe similar performances for the \textit{Wiki} and \textit{Founta} classifiers despite different class ratios in their training sets, and different performances for \textit{Wiki} and \textit{EA} classifiers despite their similar training class ratios. We also observe better performance from the \textit{CH} classifier (on the \textit{EA} test set), compared to the \textit{Wiki} or \textit{Founta} classifiers, despite the very small size of the \textit{CH} dataset. Based on previous research, we argue that cross-dataset generalization in abusive language detection is often governed by the compatibility of the definitions and sampling strategies of training and test labels rather than class sizes \cite{yin2021towards}. Instead, we explain the results presented in Table \ref{tab:generalization} in terms of implicit/explicit types of abuse and the change of vocabulary. 

\setlength{\tabcolsep}{6pt}

\begin{table}[t]
    \centering
    \small{
    \begin{tabular}{p{0.08\textwidth}p{0.08\textwidth}p{0.03\textwidth}p{0.03\textwidth}p{0.005\textwidth}p{0.03\textwidth}p{0.03\textwidth}}
 \hline 
   \multicolumn{1}{l}{\textbf{Domain}}  &\multicolumn{1}{l}{\textbf{Train Set}}   & \multicolumn{2}{c}{\textbf{AUC}}&& \multicolumn{2}{c}{\textbf{F1-score }} \\    \cline{3-7}
  & & \textit{EA} & \multicolumn{1}{c}{\textit{CH}} && \textit{EA} & \textit{CH}\\
\hline
  \multirow{2}{*}{COVID} &\textit{EA }  &0.94&0.82&&0.74&0.66\\
    &\textit{CH }  &0.86&-&&0.62&-\\
\hline
      \multirow{3}{*}{pre-COVID} &\textit{Founta }  &0.69&0.73&& 0.29&0.65\\
      
    &\textit{Wiki} &0.64&0.74 &&0.27&0.69\\
      
         & \textit{Wiki-exp} &0.58&0.71 &&0.15&0.56 \\
     \hline
 \end{tabular}
    \caption{Cross-dataset generalization on \textit{EA} (mostly implicit) and \textit{CH} (mostly explicit) datasets.
    }  
    \label{tab:generalization}
    }
\end{table}              
            
\setlength{\tabcolsep}{6pt}

\noindent \textbf{Cross-dataset generalization is better when datasets share similar vocabulary.} The classifiers trained on the \textit{EA} and \textit{CH} datasets perform better than all the classifiers trained on the pre-COVID 
datasets (\textit{Wiki} and \textit{Founta}).  
Interestingly, the performance of the \textit{CH} classifier on the \textit{EA} dataset is higher than the performance of all the general classifiers, despite the \textit{CH} dataset being very small and containing mostly explicit abuse. This observation confirms that general classifiers need to be updated to learn the new vocabulary.


\noindent\textbf{General-purpose classifiers generalize better to explicit than implicit examples in the new domain.} 
The \textit{Wiki} and \textit{Founta} classifiers, which have been exposed to large amounts of generally explicit abuse, perform well on the mostly explicit \textit{CH} dataset, but experience difficulty with the COVID-specific implicit abuse in the \textit{EA} dataset. For example, the tweet \textit{`the chinavirus is a biological attack initiated by china'} is misclassified as non-abusive. We observe that \textit{Wiki-exp} performs relatively similar to the \textit{Wiki} classifier on \textit{CH}, despite its small size (only 1,294 positive examples) but is worse than \textit{Wiki} classifier on \textit{EA}. This means that the additional 35K instances (of which, 9K are positive examples) of the \textit{Wiki} compared to the \textit{Wiki-exp}, only moderately improve the classification of the implicit examples in the new domain. This observation indicates that generalization mostly occurs between the explicit type of the pre-COVID abuse and the explicit type of the COVID-related abuse. Therefore, a general-purpose classifier should be specifically updated to learn implicit abuse in the new domain.

\section{Sensitivity to Implicit and Explicit Abuse to Explain Generalizability}

In Section \ref{sec:generalization}, we showed that when a new domain emerges, the change in vocabulary 
mostly affects the classification of implicitly expressed abuse. This observation is in line with findings by \citet{fortuna2021well}, and suggests that generalization should be evaluated on implicit and explicit abuse separately. However, due to complexities of annotation of abusive content, curating separate implicit and explicit test sets is too costly \cite{wiegand-etal-2021-implicitly}. 
Instead, we propose to adapt the Testing Concept Activation Vector (TCAV) algorithm, originally developed for image classification
\cite{kim2018interpretability}, to calculate the classifiers' sensitivity to explicit and implicit COVID-related racism, using only a small set of examples. 
Then, we show how these sensitivities can explain the generalizations observed in Table \ref{tab:generalization}. 

\subsection{TCAV background and implementation }
TCAV is a post-training interpretability method to measure how important a user-chosen concept is for a prediction, even if the concept was not directly used as a feature during the training. The concept is defined with a set of \textit{concept examples}.
To illustrate, \citet{kim2018interpretability} suggest ``stripes'' as a visual concept relevant to the class ``zebra'', and then operationally define the ``stripes'' concept by collecting examples of images containing stripes. In our language-based TCAV method, a concept is defined by a set of manually chosen textual examples. 
We collect examples from held-out subsets or other available data sources and manually annotate them for the concept of interest (for example, explicit anti-Asian abuse).
Then, we represent the concept by averaging the representations of 
the examples 
that convey that concept, similarly to how the ``stripes'' concept is represented by several images that include stripes. 

Here, we consider concepts such as COVID-19, hate speech, and anti-Asian abuse, but the approach generalizes to any concept that can be defined through a set of example texts. Using these examples, a Concept Activation Vector (CAV) is learned to represent the concept in the activation space of the classifier. Then, directional derivatives are used to calculate the sensitivity of predictions to changes in inputs towards the direction of the concept, at the neural activation layer.

We adapt the TCAV procedure for a binary RoBERTa-based classifier to measure the importance of a concept to the positive class. For any input text, $x \in \mathbb{R} ^{k\times n}$, with $k$ words in the $n$-dimensional input space, we consider the RoBERTa encoder of the classifier as $f_{emb} : \mathbb{R} ^{k\times n} \rightarrow \mathbb{R} ^{m} $, which maps the input text to its RoBERTa representation (the representation for [CLS] token), $r \in \mathbb{R}^m$. For each concept, $C$, we collect $N_C$ concept examples, 
and map them to RoBERTa representations ${r_C^j}, j=1,...,N_{C}$. To represent $C$ in the activation space, we calculate $P$ number of CAVs, $\upsilon_C^{p}$, by averaging\footnote{In the original TCAV algorithm, a linear classifier is trained to separate representations of concept examples and random examples. Then, the vector orthogonal to the decision boundary of this classifier is used as the CAV. We experimented with training a linear classifier and found that the choice of random utterances has a huge impact on the results to the point that the results are not reproducible. More stable results are obtained when CAVs are produced by averaging the RoBERTa representations.} the RoBERTa representations of $N_{\upsilon}$ randomly chosen concept examples:

\vspace{-2mm}
\begin{equation}
\small
  \upsilon_C^{p} = \frac{1}{N_{\upsilon}}\sum_{j = 1}^{{N_\upsilon}}{r_C^j} \quad p= {1,..,P}
\label{eq:CAV}
\end{equation}
 
\noindent where $N_{\upsilon}<N_C$. The \textit{conceptual sensitivity} of the positive class to the $\upsilon_C^{p}$, at input $x$ can be computed as the directional derivative $S_{C,p}(x)$:

\vspace{5pt}
$S_{C,p}(x) =
    \lim\limits_{\epsilon \to 0} \frac{h(f_{emb}(x)+\epsilon \upsilon_C^{p}) -h(f_{emb}(x))}{\epsilon}$

\begin{equation}
\quad \quad \quad = \bigtriangledown h(f_{emb}(x)).\upsilon_C^{p} 
\label{eq:sensitivity}
\end{equation}
 
\noindent where $h: \mathbb{R}^m \to \mathbb{R}$ is the function that maps the RoBERTa representation to the logit value of the positive class. In Equation \ref{eq:sensitivity}, $S_{C,p}(x)$ measures the changes in class logit, if a small vector in the direction of $C$ is added to the input example, in the RoBERTa-embedding space. 
For a set of input examples $X$, we calculate the TCAV score as the fraction of inputs for which small changes in the direction of $C$ increase the logit:

\vspace{-4mm}
\begin{equation}
    TCAV_{C,p} = \frac{| {x \in X:S_{C,p}(x)>0}|}{|X|}
\label{eq:TCAV}
\end{equation}

\noindent A TCAV score close to one indicates that for the majority of input examples the logit value increases. 
Equation \ref{eq:TCAV} defines a distribution of scores for the concept $C$; we compute the mean and standard deviation of this distribution to determine the overall 
sensitivity of the classifier to the concept $C$.

\subsection{Classifier's Sensitivity to a Concept}
\label{subsec:sensitivity}

We define each concept $C$ with $N_C =100$ manually chosen examples, and experiment with six concepts described in Table \ref{tab:concepts}.
To set a baseline, we start with a set of random examples to form a non-coherent concept. Next, we define a non-hateful COVID-related concept using random tweets with COVID-related keywords \textit{covid, corona, covid-19, pandemic}. 
For the explicit anti-Asian abuse concept, we include all 14 explicitly abusive examples from the \textit{EA} dev set and 86 explicitly abusive examples from 
\textit{CH} class. We define two implicit anti-Asian concepts with examples from \textit{EA} and \textit{CH},
to assess whether selecting the examples from two different datasets affects the sensitivities. 
We also define the generic hate concept with examples of pre-COVID general hateful utterances, not directed at Asian people or entities, from the \textit{Founta} dev set. 

\setlength{\tabcolsep}{2pt}
\begin{table}[]
    \centering
    \small{
     \begin{tabular}{p{0.98\columnwidth}}
     \hline

\textbf{Non-coherent concept:}  
random tweets collected with stop words as queries\\
\textbf{COVID-19:}  
tweets collected with words \textit{covid, corona, covid-19, pandemic} as query words\\
\textbf{Explicit anti-Asian abuse:} tweets labeled as explicit from \textit{EA} dev and \textit{CH}\\ 
\textbf{Implicit abuse (EA):} tweets labeled as implicit from \textit{EA} dev \\
\textbf{Implicit abuse (CH):} tweets labeled as implicit from \textit{CH}\\ 
\textbf{Generic hate:} tweets 
from the \textit{Hateful} class of \textit{Founta} dev\\
  \hline
 \end{tabular}
 \caption{Human-defined concepts and the sources 
 of the tweets
 used as concept examples.}
  \label{tab:concepts}}
\end{table}

\setlength{\tabcolsep}{6pt}

\begin{table*}[t]
    \centering
    \small{
    \begin{tabular}{c|c|c|c|c|c|c}
    \hline
   \multicolumn{1}{c|}{} & \multicolumn{6}{c}{\textbf{Concept}}\\
\cline{2-7}    
       \multicolumn{1}{c|}{\textbf{Classifier}} &non-coherent&COVID-19 &explicit anti-Asian&implicit (EA)&implicit (CH)&generic hate \\
       \hline
       \textit{EA}   &0.00 (0.00) &0.00 (0.00)&\textbf{0.90} (0.26)&\textbf{0.87} (0.30)&\textbf{0.70 }(0.42)&0.00 (0.00)\\
       \hline
        \textit{CH}   &0.00 (0.00) &0.00 (0.00)&-&0.35 (0.44)&-&0.21 (0.12)\\
    \hline
       \textit{Founta}   & 0.00 (0.02) &0.00 (0.01) &\textbf{0.92} (0.22) &0.00 (0.06)&0.19 (0.32)&\textbf{0.60 }(0.44)\\
       \hline
       \textit{Wiki}  &0.00 (0.03) &0.00 (0.05)&\textbf{0.96 }(0.16)&0.00 (0.03) &0.32 (0.44)& \textbf{0.75 }(0.41)\\
     \hline
        \textit{Wiki-exp}   &0.00 (0.05) &0.00 (0.07)&\textbf{0.78} (0.12)&0.00 (0.02)&0.00 (0.05)&\textbf{0.59} (0.40)\\
     
     \hline
    \end{tabular}
    \caption{Means and standard deviations of TCAV score distributions for the positive class of the five classifiers with respect to six human-defined concepts. Scores statistically significantly  different  from random are in bold. }
    \label{tab:sensitivity}
    }
\end{table*}

We calculate $P=1000$ CAVs for each concept, where each CAV is the average of $N_\upsilon =5$ randomly chosen concept examples. We use 2000 random tweets collected with stopwords as input examples $X$ (see Equation~\ref{eq:TCAV}).\footnote{Unlike the original TCAV algorithm, we do not restrict the input examples to the target class. In our experiments, we observed that, for this binary classification set-up, the choice of input examples has little impact on the TCAV scores. Intuitively, we assess whether adding the concept vector to a random input would increase the likelihood of it being assigned to the positive class.} Table \ref{tab:sensitivity} presents the means and standard deviations of the TCAV score distributions for the classifiers trained on \textit{Wiki}, \textit{Founta}, \textit{EA}, and \textit{CH} datasets, respectively. 
First, we observe that all TCAV scores calculated for a  random, non-coherent set of examples are zero; i.e., as expected, the TCAV scores do not indicate any association between a non-coherent concept and the positive class. 
Also, as expected, none of the classifiers associate the non-hateful COVID-related concept to the positive class. Note that a zero TCAV score can be due to the absence of that concept in the training data (e.g., the COVID concept for the \textit{Wiki} and \textit{Founta} classifiers), insignificance of the topic for predicting the positive label (e.g., the COVID concept for the \textit{EA} classifier), or the lack of coherence among the concept examples (such as the concept defined by random examples). A TCAV score close to 1, on the other hand, indicates the importance of a concept for positive prediction. These observations set a solid baseline for interpreting the TCAV scores, calculated for other concepts. Here we ask whether the generated TCAV scores can explain the generalization performances observed in Table~\ref{tab:generalization}.

We consider a classifier to be sensitive to a concept if its average TCAV score is significantly different (according to the t-test with p < 0.001) from the average TCAV score of a non-coherent random concept.  First, we observe that 
the general classifiers 
are only sensitive to the explicit type of COVID-related abusive language. This confirms that 
the classifiers generalize better to the explicit type of an emerging domain of abusive language. We also note that \textit{Wiki-exp}, 
is sensitive to the explicit anti-Asian concept. 

Second, the classifier trained with mostly explicit COVID-related data (\textit{CH}) is not sensitive to the implicit abuse concept.\footnote{We do not measure the sensitivity of this classifier to the explicit anti-Asian and implicit CH concepts, since their concept examples are included in the training set of the classifier.} 
The only classifier that is sensitive to the explicit and both implicit COVID-related abusive concepts is the \textit{EA} classifier. Classifiers trained on the COVID datasets are also not sensitive to the generic hate concept, which encompasses a much broader range of target groups. Overall, these findings stress the importance of including implicitly abusive examples in the training data for better generalizability within and across domains.

\section{Degree of Explicitness}
\label{sec:DoE}

Here, we suggest that implicit examples are more informative (less redundant) for updating a general classifier and provide a quantitative metric to guide the data augmentation process. We extend the TCAV methodology to estimate the \textit{Degree of Explicitness} or \textit{DoE} of an utterance. We showed that the average TCAV score of the positive class for the explicit concept is close to 1. DoE is based on the idea that adding one example to an explicit concept will not affect its average TCAV score (i.e., it will still be close to 1), if the added example is explicitly abusive. However, adding an implicit example presumably will change the direction of all CAVs and reduce the sensitivity of the classifier to this modified concept. 
Here, we modify Equation \ref{eq:CAV} and calculate each CAV by averaging the RoBERTa representations of $N_{\upsilon}-1$ explicit concept examples, and the new utterance for which we want the degree of explicitness, $x_{new}$, with representation $r_{new}$. Thus, 
\vspace{-4mm}
\begin{equation*}
    \upsilon_{new}^{p} = \frac{1}{N_{\upsilon}}(\sum_{j = 1}^{{N_\upsilon-1}}{r_C^j+r_{new}}), \quad p= {1,..,P}
\end{equation*}
\vspace{-4mm}

\noindent We then calculate the average TCAV score for each $x_{new}$ as its DoE score. If the new utterance, $x_{new}$, is explicitly abusive, $\upsilon_{new}^{p}$ will represent an explicit concept, and the average TCAV score, i.e., $mean(TCAV_{C,p})$ will remain close to 1. However, the less explicit the new example is, the more $\upsilon_{new}^{p}$ will diverge from representations of explicit abuse, and the average score will drop. We use $N_{\upsilon}=3$ in the following experiments.

\begin{figure}[t]
\centering
\includegraphics[width = \columnwidth]{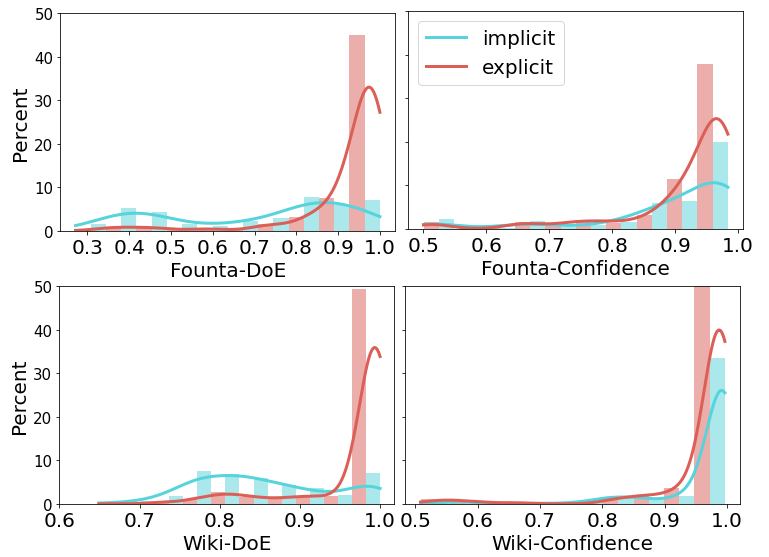}
\caption{Comparison of classification confidence and DoE score for distinguishing between implicit and explicit abusive utterances.}   
\label{fig:prob-DoE}
\end{figure}

\noindent \textbf{DoE analysis on COVID-related abusive data:} We validate the utility of DoE in terms of separating implicit and explicit abusive examples. For the \textit{Wiki} and \textit{Founta} classifiers, we calculate the DoE score of the implicit and explicit examples from \textit{CH} and the \textit{EA} dev set (described in Section \ref{sec:generalization}), excluding the examples used to define the \textit{Explicit anti-Asian abuse} concept. Given that low classification confidence could indicate that the model struggles to predict an example correctly, one might expect that implicit examples are classified with less classification confidence than explicit examples. 
Figure \ref{fig:prob-DoE} shows the comparison of DoE with classification confidence in distinguishing between implicit and explicit examples. We observe that for both classifiers, the distribution of DoE scores of implicit examples is different from the distribution of DoE scores of explicit examples, but the distributions of their classification confidences are indistinguishable. Therefore, we conclude that DoE is more effective at separating implicit abuse from explicit abuse than classification confidence. We further analyze DoE scores for the positive and negative classes separately in Appendix \ref{sec:appendix-B}.

\section{Data Augmentation with DoE score}
\label{sec:augmentation}
We now use the DoE score to direct data augmentation. We consider a scenario where a general classifier should be re-trained with an augmented dataset to include emerging types of abusive language. As we showed, general classifiers are already sensitive to explicit abuse. Therefore, we hypothesize that implicit examples are more beneficial for updating the classifier. Here, we describe a novel DoE-based augmentation approach and contrast it with the conventional process of choosing augmentation examples based on the classification confidence \cite{zhu2008learning,chen2019active}.  

We consider the general \textit{Wiki} classifier.
Our goal is to find a small but sufficient portion of the \textit{EA} train set to augment the original \textit{Wiki} train set, so that the classifier is able to handle 
COVID-related anti-Asian hate speech. 
We calculate the DoE and confidence scores for all the examples in the \textit{EA} train set and add the $N$ examples with the lowest scores to the original \textit{Wiki} train set.
We vary $N$ from 1K to 6K, with a 1K step. 
After the augmentation data size reaches 6K, the classifier performance on the original \textit{Wiki} test set drops substantially for both techniques. 
Also, note that as the size of the augmentation dataset increases, the two methods converge to the same performance.

\subsection{Results}
\label{subsec:aug-clasifier}

\begin{figure}[t]
\centering
\includegraphics[width = \columnwidth]{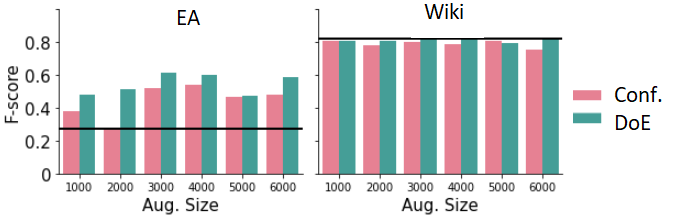}
\caption{F1-score of the augmented \textit{Wiki} classifier on the \textit{EA} and \textit{Wiki} test sets. Solid lines show the baseline.}
\label{fig:augmentation}
\end{figure}

Figure \ref{fig:augmentation} shows the F1-score of the classifiers updated using the DoE and confidence-based augmentation methods on the original test set (\textit{Wiki}) and the new test set (\textit{EA}) for different augmentation sizes. (Precision and recall figures are provided in Appendix \ref{sec:appendix-C}.) Since only \textit{EA} is used for augmentation, we evaluate the classifiers on this dataset to find the optimum size for the augmented training set and only evaluate the best performing classifiers on \textit{CH}.   
We expect that an efficient augmentation should maintain the performance on \textit{Wiki} and reach acceptable results on \textit{EA} test set. 

\noindent\textbf{DoE is better at learning the new type of abuse:} On the \textit{EA} dataset, DoE achieves better results than the confidence-based augmentation method for all augmentation sizes, except for N= 5K, where the performances of the two methods are comparable.

\noindent\textbf{DoE is better at maintaining performance on the original dataset:}
DoE outperforms the confidence-based method 
on the \textit{Wiki} dataset. For all augmentation sizes, the performance of the DoE-augmented classifier on this class stays within 2\% of the baseline (the F1-score of the classifier trained just on the \textit{Wiki} data), whereas for the confidence-based augmentation, we observe up to 6\% drop depending on the size of the added data. 

\noindent \textbf{DoE is better overall:} 
Table \ref{tab:aug-results} presents the best results achieved by the two augmentation methods on the \textit{EA} test set: AUC score of 0.81 for the DoE-based augmentation obtained with 3K added examples, and AUC score of 0.69 for the confidence-based augmentation obtained with 4K added examples. For comparison, we also show the baseline results for the original \textit{Wiki} classifier and the classifier trained on the combined \textit{Wiki} and full \textit{EA} train sets. Although we did not optimize the augmentation for the \textit{CH} dataset, our evaluation shows that DoE performs favourably on this dataset, as well. We conclude that the new DoE-based augmentation method maintains the classification performance on the original dataset, while outperforming the other method on the new data.

We also qualitatively assess the classifier's output before and after data augmentation with DoE. 
While explicitly abusive utterances (e.g., ``f*ck you china and your chinese virus'') are often correctly classified both before and after re-training, many implicitly abusive examples (e.g., ``it is not covid 19 but wuhanvirus'') are handled correctly by the classifier only after re-training.

\begin{table}[t]
    \centering
    \small{
    \begin{tabular}{p{0.05\textwidth} p{0.08\textwidth}p{0.02\textwidth}p{0.02\textwidth}p{0.02\textwidth}p{0.0001\textwidth}p{0.02\textwidth}p{0.02\textwidth}p{0.02\textwidth}}

& & \multicolumn{3}{c}{F1-score}&&\multicolumn{3}{c}{AUC}\\
      \hline  
    \textbf{Method}&\textbf{Aug. set}& \textit{\textbf{EA}} & \textit{\textbf{CH}}  & \textit{\textbf{Wiki}}&&\textit{\textbf{EA}} & \textit{\textbf{CH}}  & \textit{\textbf{Wiki}}\\
      \hline  
DoE&3K \textit{EA} &	\textbf{0.61}&\textbf{0.73}&\textbf{0.82}&&\textbf{0.81}&\textbf{0.78}&\textbf{0.96}\\
Conf.& 4K \textit{EA}&	0.54&	0.71&0.79&&0.69&0.75&0.94\\
Merging&\textit{EA}&0.58&0.72&0.78&&0.72&0.75&0.94\\
\hline
baseline& -&0.27&0.69&0.82&&0.64&0.74&0.96\\

    \hline
 \end{tabular}
    \caption{AUC and F1-scores for the best
    performing
    classifiers updated with various augmentation methods, as well as the original \textit{Wiki} classifier as baseline. }
    \label{tab:aug-results}
    }
\end{table}

\section{Related Work} 

Generalizability has been an active research area in NLP  \cite{ettinger-etal-2017-towards,hendrycks-etal-2020-pretrained}. In a recent review, \citet{yin2021towards} discussed the challenges for building generalizable hate speech detection systems and recommended possible future directions, including improving data quality and reducing overfitting through transfer learning.
Several studies evaluated generalizability in abuse detection through cross-dataset evaluation \cite{swamy-etal-2019-studying,wiegand2019detection}, direct dataset analysis \cite{fortuna2020toxic} or topic modeling on the training data \cite{nejadgholi-kiritchenko-2020-cross}.
\citet{fortuna2021well} showed that the lack of generalizability is rooted in the imbalances between 
implicit and explicit examples in training data. 

The distinction between explicit and implicit abuse has been recognized as an important factor in abuse detection \cite{waseem-etal-2017-understanding,caselli-etal-2020-feel}. 
\citet{wiegand2019detection} showed that lexicon-based sampling strategies fail to collect implicit abuse and most of the annotated datasets are overwhelmed with explicit examples. \citet{breitfeller-etal-2019-finding} showed that inter-annotation agreement is low when labeling the implicit abuse utterances, as sometimes specific knowledge is required in order to understand the implicit statements. For better detection of implicitly stated abuse,  large annotated  datasets  with hierarchical annotations  are  needed \cite{sap2020socialbiasframes}, so  that  automatic  detection  systems  can learn from a  wide variety of such training examples. \citet{field2020unsupervised} proposed propensity matching and adversarial learning to force the model to focus on signs of implicit bias. 
\citet{wiegand-etal-2021-implicitly} created a novel dataset for studying implicit abuse and presented a range of linguistic features for contrastive analysis of abusive content. We define explicitness as obvious rudeness and hateful language regardless of the social context and introduce a quantitative measure of explicitness from a trained classifier's point of view.

Data augmentation has been used to improve the robustness of abuse detection classifiers. To mitigate biases towards specific terms (e.g., identity terms), one strategy is to add benign examples containing the biased terms to the training data \citep{dixon2018measuring,badjatiya2019stereotypical}. 
Other works combined multiple datasets to achieve better generalizations, using a set of probing instances \cite{han-tsvetkov-2020-fortifying}, multi-task training \cite{waseem2018bridging}, and domain adaptation \cite{karan-snajder-2018-cross}. 
In contrast to these works, we take an interpretability-based approach and guide the data collection process by mapping the new data on the implicit vs.\@ explicit spectrum.

\section{Conclusion} 

As real-world data evolves, we would like to be able to query a trained model to 
determine whether it generalizes to the new data, without the need for a large, annotated test set. We adopted the TCAV algorithm to quantify the sensitivity of text classifiers to human-chosen concepts, defined with a small set of examples. 
We used this technique to compare the generalizations of abusive language classifiers, trained with pre-pandemic data, to explicit and implicit COVID-related anti-Asian racism.

We then proposed a sensitivity-based data augmentation approach, to improve generalizability to emerging categories. We showed that in the case of abuse detection, the most informative examples are implicitly abusive utterances from the new category. Our approach collects implicit augmentation examples and achieves higher generalization to the new category compared to confidence-based sampling. Strategies for choosing the optimal set of concept examples should be explored in the future. 

While we examined abusive language detection as a case study, similar techniques can be applied to different NLP applications. 
For example, the TCAV method could be used to measure the sensitivity of a sentiment analysis system to a new product, or a stance detection algorithm's sensitivity to an important new societal issue. As language evolves, methods of monitoring and explaining classifier behaviour over time will be essential.

\section*{Ethical Considerations}

Content moderation is a critical application with potential of significant benefits, but also harms to human well-being. Therefore, ethics-related issues in content moderation have been actively studied in NLP and other disciplines \cite{vidgen-etal-2019-challenges,wiegand2019detection,kiritchenko2020confronting,vidgen2020directions}. These include sampling and annotation biases in data collection, algorithmic bias amplification, user privacy, system safety and security, and human control of technology, among others. Our work aims to address the aspects of system safety and fairness by adapting the model to newly emerged or not previously covered types of online abuse, often directed against marginalized communities. 
We employ existing datasets (with all their limitations) and use them only for illustration purposes and preliminary evaluation of the proposed methodology.
When deploying the technology care should be taken to adequately address other ethics-related issues.

\bibliography{main}
\bibliographystyle{acl_natbib}

\clearpage 

\appendix

\section{Model Specifications}
\label{sec:appendix-A}

All of our models are binary RoBERTa-based classifiers trained with the default settings of the Trainer module from the Huggingface library\footnote{\url{https://huggingface.co/transformers/main_classes/trainer.html}} for 3 training epochs, on a Tesla V100-SXM2 GPU machine, batch size of 16, warm-up steps of 500 and weight decay of 0.01. We use Roberta-base model, which includes 12 layers, 768 hidden nodes, 12 head nodes, 125M parameters, and add a linear layer with two nodes for binary classification. Training these classifiers takes several hours depending on the size of the training dataset.




\section{Additional Results for Cross-Dataset Generalization}
\label{sec:appendix-A1}

In table \ref{tab:generalization_additional}, we present additional metrics for the generalizibility experiments described in Section~\ref{sec:generalization}. Besides the commonly used metrics, precision and recall, we measure averaged precision score to count for potential threshold adjustments. Averaged precision score summarizes a precision-recall curve as the weighted mean of precisions at each threshold, weighted by the increase in recall from the previous threshold. The results are consistent with AUC and F1-scores reported in Table \ref{tab:generalization}.

\setcounter{table}{0}
\renewcommand\thetable{B.\arabic{table}}

\setlength{\tabcolsep}{5pt}
\begin{table}[hbt!]
    \centering
    \small{
    \begin{tabular}{p{0.07\textwidth}p{0.03\textwidth}p{0.03\textwidth}p{0.001\textwidth}p{0.03\textwidth}p{0.03\textwidth}p{0.001\textwidth}p{0.03\textwidth}p{0.03\textwidth}}
 
 \hline
  \multicolumn{1}{c}{\textbf{Train Set}}& \multicolumn{2}{c}{\textbf{Precision}}&& \multicolumn{2}{c}{\textbf{Recall }}&& \multicolumn{2}{c}{\textbf{Ave. Prec. }} \\  
  \cline{2-9}
    \multicolumn{1}{c}{\textbf{}}& \multicolumn{1}{c}{\textit{EA}} & \multicolumn{1}{c}{\textit{CH}} && \multicolumn{1}{c}{\textit{EA}} & \multicolumn{1}{c}{\textit{CH}}&& \multicolumn{1}{c}{\textit{EA}} & \multicolumn{1}{c}{\textit{CH}}\\
  \hline
  \textit{EA }  &0.72&0.77&&0.73&0.58&&0.80 &0.80\\
    \textit{CH }  &0.58&-&&0.66&-&&0.64 &-\\
    \hline

     \textit{Founta }  &0.46&0.57&& 0.23&0.73&&0.35&0.65\\
      
     \textit{Wiki} &0.39&0.61&&0.21&0.78&&0.31 &0.66\\
      
          \textit{Wiki-exp} &0.37&0.64&&0.10&0.51 &&0.26 &0.64 \\
      \hline
 \end{tabular}
    \caption{Additional metrics for cross-dataset generalization results presented in Table \ref{tab:generalization}.
    }  
    \label{tab:generalization_additional}
    }
\end{table}
\setlength{\tabcolsep}{6pt}

\section{DoE Analysis on the \textit{EA} Train Set}
\label{sec:appendix-B}

With the DoE score, we want to distinguish between implicit and explicit examples of abuse. However, when used for data selection, the true labels of the selected examples are not available. We investigate what low DoE scores mean in terms of `being challenging to classify'. With both \textit{Founta} and \textit{Wiki} classifiers, we calculate the DoE score for all instances of the \textit{EA} train set, sort the negative and positive examples separately based on DoE and look at the classification accuracies in bins of size 100 of sorted DoEs. 
Figure \ref{fig:doe-conf} shows that low DoE examples are correctly classified if negative and misclassified if positive (implicit abuse). In contrast, high DoE examples are misclassified if negative and correctly classified if positive (explicit abuse). 

\setcounter{figure}{0}
\renewcommand\thefigure{C.\arabic{figure}}

\begin{figure}[h]
\centering
\includegraphics[trim = 0cm 0cm 0cm 0cm, clip,width = 0.8\columnwidth]{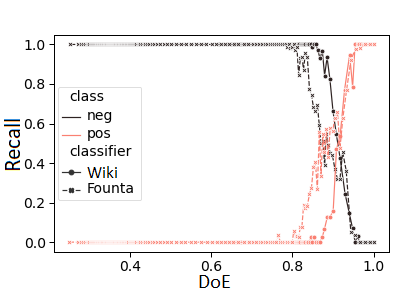}
\caption{Recall per class for varying DoE scores on the \textit{EA} train set} 
\label{fig:doe-conf}
\end{figure}

\section{Comparing DoE and Confidence-Based Augmentation Using Precision and Recall}
\label{sec:appendix-C}

In Section~\ref{sec:augmentation}, we compare the classifiers updated with DoE and confidence-based methods using classification F1-score. Here, we provide a more fine-grained analysis based on recall and precision.

Figure \ref{fig:augmentation_per_rec}  shows the recall and precision of the updated classifiers on the \textit{EA} dataset. This figure indicates that the classifiers updated with DoE are much more successful in recognizing abusive utterances than the classifiers updated with confidence, but misclassify more non-abusive sentences, which results in substantially higher recall scores, but slightly lower precision scores. Note that in computer-assisted content moderation, recall is more important than precision, since automatically flagged posts are assessed by human moderators to make the final decision.

We argue that the higher recall and lower precision of classifiers updated with DoE is due to the discrepancies in the definitions of the negative classes for the \textit{Wiki} and \textit{EA} datasets. Previous work has commented on the difficulty of aligning annotations of  \textit{abusive}, \textit{offensive}, \textit{hateful}, and \textit{toxic} speech across different datasets \cite{swamy-etal-2019-studying, kolhatkar2019sfu, fortuna2021well}.
 Here, we also observe that the definitions of positive (abusive) and negative classes differ significantly between the generalized and COVID-related data. 
 In the \textit{Wiki} and \textit{Founta} datasets, the positive class encompasses a wide range of offensive language, while in the \textit{EA} and \textit{CH} datasets, the positive class is restricted to hate speech and other more intense cases of expressed negativity. 
Further, the negative class in \textit{Wiki} and \textit{Founta} datasets comprise non-abusive, neutral, or friendly instances while in the \textit{EA} and \textit{CH} datasets the negative class may also include rude and offensive texts as long as they do not constitute hate speech against Asian people or entities. 

In Appendix \ref{sec:appendix-B}, we observe that low DoE examples are correctly classified if negative and misclassified if positive (implicit abuse). In contrast, high DoE examples are misclassified if negative and correctly classified if positive (explicit abuse). We use this observation to explain higher recall of the confidence-based method in comparison with the DoE-based method for the \textit{EA-negative} class. As mentioned before, while \textit{EA-positive} fits under the definition of `toxicity' in  \textit{Wiki-positive}, the definition of \textit{EA-negative} is inconsistent with the definition of \textit{Wiki-negative}. 
In other words, DoE tends to choose negative examples that the \textit{Wiki} classifier already recognizes as negative, whereas the confidence-based data augmentation selects negative examples that are unknown to the classifier. Therefore, the classifier augmented with low confidence scores adapts better to the new definition of negative examples than the classifier updated with low DoE scores. In a real-life scenario, we do not expect the definition of the negative class to change over time, so precision for DoE-base augmentation should not suffer.

\setcounter{figure}{0}
\renewcommand\thefigure{D.\arabic{figure}}
\begin{figure}[h]
\centering
\includegraphics[width = 0.9\columnwidth]{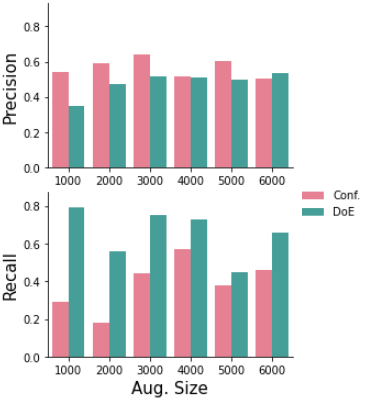}
\caption{Precision and recall of the augmented \textit{Wiki} classifier on the \textit{EA} test set.}
\label{fig:augmentation_per_rec}
\end{figure}

\end{document}